\documentclass[sigconf]{acmart}
\usepackage{multirow}
\usepackage{rotating}
\usepackage{makecell}
\usepackage{enumitem}
\usepackage{tikz}
\usepackage[table]{xcolor}
\usetikzlibrary{arrows.meta, positioning}
\usepackage{booktabs, amsmath}
\newcommand{\pz}{\phantom{$+$}}

\AtBeginDocument{%
  }

\begin{document}

\title{MERCI Cards: An LLM Evaluation and Deployment Framework for High-Stakes Domains}

\author{Aparna Komarla}
\authornote{Aparna is the founder of Redo.io (\url{https://getredo.io}), a legal AI nonprofit building technology solutions for resentencing implementation in California; a Responsible AI Fellow at the Duke University School of Law; and a graduate student in Symbolic Systems at Stanford University.}
\email{akomarla@stanford.edu}
\affiliation{
  \institution{Redo.io (https://getredo.io)}
  \city{Berkeley}
  \state{California}
  \country{USA}
}
\affiliation{
  \institution{Stanford University}
  \city{Palo Alto}
  \state{California}
  \country{USA}
}

\author{Annalisa Szymanski}
\authornote{Annalisa Szymanski is a PhD candidate in the Human-Computer Interaction (HCI) Lab at the University of Notre Dame}
\email{aszyman2@nd.edu}
\affiliation{
  \institution{University of Notre Dame}
  \city{Notre Dame, Indiana}
  \country{USA}
}


\begin{abstract}
As LLMs are increasingly deployed in high-stakes professional workflows, engineers and researchers require principled protocols to systematically track, monitor, and improve model performance across deployment cycles. We present a mathematical framework for iterative LLM evaluation and deployment, and demonstrate its application to AI systems used in criminal justice. Our framework formalizes LLM integration in high-stakes, high-risk, and resource-constrained domains across model selection, rubric design, evaluations and deployment via a weighted multi-objective optimization. We demonstrate that MERCI Cards can guide improvements of the system across deployment iterations, direct developer attention toward under-performing areas, and focus user attention on validation and error-correction in the LLM's outputs.
\end{abstract}


%%  \includegraphics[width=\textwidth]{sampleteaser}
%%  \caption{Seattle Mariners at Spring Training, 2010.}
%%  \Description{Enjoying the baseball game from the third-base
%%  seats. Ichiro Suzuki preparing to bat.}

\maketitle

\section{Introduction}

Deploying LLMs in professional workflows demands ongoing judgment: when are outputs reliable enough, when should human reviewers be involved, and when does a newer model justify switching costs? LLM judges offer scalable evaluation but are susceptible to self-preference bias and rubric ceiling effects \cite{llm_judge_self_bias_1, llm_judge_self_bias_2, bavaresco_llm_judge_2025, gu2024_llmjudge, yancey_2023_rating}. Human evaluators offer domain expertise but are costly and may encode interpretive biases that are difficult to tease out \cite{chen-etal-2024-humans, shankar2025}. Few established protocols exist that govern how to combine evaluator types, select general-purpose AI models, and make principled tradeoffs between quality, cost, and human involvement. The absence of guidance for building effective systems, and transparency about deployed systems, has made some practitioners skeptical of AI integrations overall, and sometimes opt to avoid AI adoption \cite{abandon_ai_dev}.

This information gap is particularly consequential in domains where access to technology is unequal and the stakes of error are high. For instance, the cost of repeating an evaluation with human experts is amplified in organizations with funding constraints for core operations. But without a disciplined evaluation framework, it is difficult to know whether a given system is ready for deployment and how to improve it.

We propose a mathematical framework for LLM evaluation and deployment: Model-Evaluation-Rubric-Iterate, MERCI Cards for short---a framework for AI deployment and integration for transparency and observability. MERCI Cards are inspired by Model Cards among other documentation tools for machine learning and automated decision making systems \cite{model_cards_ml}. The framework covers model selection, evaluation rubric design, and human-LLM evaluation feedback across multiple iterations. Section~\ref{sec:framework} presents the formal design and MERCI Card and Section~\ref{sec:application} introduces an application of the framework to the RJA and reports findings from real-world deployments.

\subsection{Research Questions}
Our work addresses three questions: \textbf{RQ1.} How should model selection, evaluation, and iteration be formalized for LLM integration in high-stakes, high-risk and resource-constrained settings? \textbf{RQ2.} How much does protocol for model selection and evaluation influence the accuracy of system outputs, and the cost and time required to generate them? \textbf{RQ3.} What system design choices---model family, reasoning level, context, and rubric---most influence system performance?

\section{Framework Design}\label{sec:framework}

AI-assisted generation encodes choices about the model, the prompt, and
the supporting context. Evaluation encodes choices about the rubric, the
evaluating model, and the degree and timing of expert involvement. Each
choice may be revised in light of the outcomes of prior evaluation,
producing an iterative pipeline. We formalize this pipeline for domains
where quality, cost, latency, and output integrity all matter, so that
each deployment cycle produces an output paired with the diagnostics
needed to decide whether to release it, how to improve it, and whether a
new model release warrants adoption.

The framework is agnostic to the underlying system. The output may be an answer in a question-answering system, a synthesized report, a document summary, a drafted communication, or any other model-produced artifact for an end-user. It makes three commitments that distinguish it from ad hoc evaluation practice. Cost and latency are selection criteria rather than post hoc considerations, which matters
where repeating an expert evaluation is itself a binding constraint.
Human and machine evaluation are kept as separate signals with separate
roles, so scarce expert attention can be directed where it has the most
value. And every configuration choice and constraint violation is
recorded on a versioned artifact, the MERCI card, issued once per
iteration.

We refer throughout to the \emph{deploying team}: whoever is accountable
for putting the system in front of users, whether that is a product
owner, an engineering team, a research group, or an organization
adopting a tool built elsewhere. The framework does not tell this team
what to choose. It fixes the structure of the pipeline and the set of
choices that structure exposes, and it requires that each choice, and the
reasoning behind it, be written down where a later reader can find it.
What counts as an acceptable rubric, a trustworthy reviewer panel, or a
releasable score is a judgment about the domain, and the people who
understand the domain are the ones positioned to make it. What the
framework supplies is the obligation to make those judgments explicit
rather than incidental.

\subsection{Setup}\label{sec:setup}

\paragraph{Task and outputs.}
Let $\mathcal{T}$ be a generation task. At each iteration $\tau$ the task
is run on $N$ inputs, producing task results
$\mathit{TR}_1,\ldots,\mathit{TR}_N$. A task result is the user-facing
artifact and the unit of evaluation throughout. The inputs are assumed
similar enough that one configuration is expected to serve all of them,
which is what licenses averaging scores across them.

\paragraph{Iterations.}
Iterations are indexed $\tau = 0, 1, 2,$ and so on. A configuration is the
tuple $(G, P, C, R, W)$: generation model, prompt, context, rubric, and
rubric weights. Every input below belongs to some iteration; we suppress
the superscript $(\tau)$ for readability except where two iterations are
compared.

\paragraph{Rubric and scores.}
Quality is assessed along a total of $D$ rubric dimensions
$R = \{d_1,\ldots,d_D\}$. Each dimension is scored on an ordinal scale
such as $[0,1]$, where $1$ denotes full agreement with the ideal response
and $0$ none. The framework fixes the range and the direction; the
deploying team sets the granularity and records it on the
card.\footnote{Scores can be continuous, discrete or categorical (e.g.,
full credit mapped to 1, partial credit mapped to 0.5 and no credit
mapped to 0) all resulting in outputs on a $[0,1]$ scale.}

\paragraph{Evaluators and scores.}
Two evaluator types may be activated. The LLM judge $J$ scores each task
result $\mathit{TR}$ a total of $\ell$ times, since judge scores are
stochastic and their disagreements can be a diagnostic. A total of $V$
human reviewers are recruited to score a task result once. Aggregating over task results and over runs or reviewers gives two scores per dimension: $J_d$, the aggregate judge score on dimension $d$, and $H_d$, the aggregate human
score on dimension $d$. These combine into the \emph{quality vector}
$Q = \left[\, Q_1, \ldots, Q_D \,\right]$ through another aggregation function
the deploying team chooses and records on the card:

\begin{equation}\label{eq:agg}
Q_d = \mathrm{agg}\!\left(J_d,\, H_d\right)
\end{equation}

\noindent The choice of $\mathrm{agg}$ encodes an opinion on whose
judgment counts, and different organizations will reasonably take different
positions. A team using both evaluator types might take the mean,
$\mathrm{agg} = \tfrac{1}{2}(J_d + H_d)$, treating the two as equally
informative about the performance of the system; a weighted mean
$\lambda J_d + (1-\lambda) H_d$ to let expert judgment dominate without
discarding the judge signal. A team may choose to admit $H_d$ only if they pass a reliability threshold explained in Step~F, so $\mathrm{agg}$ may reduce to $J_d$ at an iteration if such a threshold is not met.

$Q$ is the feedback signal: it drives the revisions in Steps~A--C, the
model selection in Step~D, and the release decision in Step~H.

\paragraph{Cost and latency.}
Mean cost $c$ and mean latency $t$ per task result are recorded
alongside quality but are not part of $Q$. They enter model selection in
Step~D.

\paragraph{Models.}
A model $m$ is characterized by its family $\mathcal{F}(m)$ and its
reasoning level $\rho(m) \in \{0,1,2,3,4\}$. Families expose this control
under different names and native scales, so the deploying team states the
mapping from a family's settings onto $\{0,\ldots,4\}$ on the MERCI card.
For example, where $\mathcal{F}(m)$ is Anthropic, the native setting
takes values in $\{\text{none},\ \text{adaptive},\ \text{max}\}$, and the
team declares which normalized level each corresponds to. The mapping is
a stated assumption rather than a measurement, which is why it is
recorded: a reader who disagrees with it can see exactly what was assumed
and recompute accordingly.

\begin{table}[t]
\centering
\footnotesize
\caption{Notation.}
\label{tab:notation}
\setlength{\tabcolsep}{4pt}
\renewcommand{\arraystretch}{1.15}
\begin{tabular}{@{}l p{5.75cm}@{}}
\toprule
\textbf{Symbol} & \textbf{Meaning} \\
\midrule
$\mathcal{T}$, $\mathit{TR}_n$, $N$ & Task; $n$-th task result; number
  of results \\
$\tau$ & Iteration index (suppressed when unambiguous) \\
$E$ & Domain expert input \\
\addlinespace[3pt]
$G$, $P$, $C$, $R$, $W$ & Configuration: model, prompt, context,
  rubric, weights \\
$D$, $d$ & Number of rubric dimensions; dimension index \\
\addlinespace[3pt]
$J$, $\ell$ & LLM judge; judge runs per task result \\
$V$ & Number of human reviewers \\
$J_d$, $H_d$ & Mean judge and human score on dimension $d$ \\
$\mathrm{agg}$, $\lambda$ & Aggregation of $J_d$ and $H_d$ into $Q_d$;
  its weight, where weighted \\
$Q_d$, $Q$ & Quality score on $d$; quality vector
  $[\,Q_1,\ldots,Q_D\,]$ \\
$\phi$, $\delta_\phi$ & Inter-rater reliability and its threshold \\
\addlinespace[3pt]
$\mathcal{M}$, $m$ & Candidate model set; a candidate \\
$\mathcal{F}(m)$, $\rho(m)$ & Model family; reasoning level $0$--$4$ \\
$c_m$, $t_m$ & Mean cost and latency per task result \\
$w_d$, $w_c$, $w_t$ & Dimension, cost, and latency weights \\
$B_c$, $B_t$ & Cost and latency budgets, where used \\
$Q_d(m)$ & Quality score for candidate $m$ during selection \\
$\hat{S}_m$ & Composite selection score \\
\addlinespace[3pt]
$\delta_J$, $\delta_H$ & Overall judge and human release thresholds \\
$\delta_{J,d}$, $\delta_{H,d}$ & Per-dimension release floors \\
\bottomrule
\end{tabular}
\end{table}

\begin{table}[t]
\centering
\footnotesize
\caption{Choices the framework leaves to the deploying team. The
framework requires that each be recorded on the MERCI card with its
rationale; it does not prescribe the value.}
\label{tab:decisions}
\setlength{\tabcolsep}{4pt}
\renewcommand{\arraystretch}{1.15}
\begin{tabular}{@{}p{4.3cm} l l@{}}
\toprule
\textbf{Decision} & \textbf{Symbol} & \textbf{Step} \\
\midrule
Scoring granularity per dimension & --- & Setup \\
Aggregation of judge and human scores & $\mathrm{agg}$, $\lambda$ & Setup \\
Rubric dimensions & $R$ & A \\
Dimension weights & $W$ & A \\
Whether context is supplied, and whether it is withheld during
  selection & $C$ & C \\
Candidate model set & $\mathcal{M}$ & D \\
Cost and latency weights, or hard budgets & $w_c$, $w_t$,
  $B_c$, $B_t$ & D \\
Judge model, and whether to accept a constraint violation & $J$ & E \\
Judge runs per task result & $\ell$ & E \\
Whether to use human review, and panel size & $V$ & F \\
Reliability metric and threshold & $\phi$, $\delta_\phi$ & F \\
Release thresholds and per-dimension floors & $\delta_J$, $\delta_H$,
  $\delta_{\cdot,d}$ & H \\
\bottomrule
\end{tabular}
\end{table}

\subsection{Objective}\label{sec:objective}

The pipeline seeks a configuration that maximizes quality while keeping
cost and latency low, with all weights non-negative:

\begin{equation}\label{eq:objective}
    \max_{G, P, C, R, W} \;\;
    \sum_{d=1}^{D} w_d\, Q_d \;-\; w_c\, c \;-\; w_t\, t
\end{equation}

\noindent Equation~\eqref{eq:objective} is not solved jointly: the search space is combinatorial,
the quality term is observable only by running the pipeline, and the
prompt, context, and rubric are artifacts rather than optimizable
parameters. The pipeline instead improves one input at a time. The
expert revises $R$, $W$, $P$, and $C$ using the previous iteration's
quality vector (Steps~A--C), and the model coordinate is re-optimized by
an explicit search over a finite candidate set with the rest held
constant (Step~D). The composite score $\hat{S}_m$ of Step~D is
Equation~\eqref{eq:objective} restricted to the $G$ coordinate, with
$Q_d(m)$ the same quality score measured for a single candidate.
Iteration therefore produces a documented improvement record rather than
a model convergence, and Step~H defines what counts as good enough to
stop.

The weights $w_d$, $w_c$, and $w_t$ are set by the deploying team and are
where domain priorities enter the arithmetic. A team that treats latency
as a convenience and a team that treats it as a determinant of whether
the tool gets used at all will set $w_t$ very differently, and both are
defensible given their circumstances. Recording the weights alongside the
outcome is what lets a later reader distinguish a model that won on
quality from one that won on price.

%% =====================================================================
\subsection{Pipeline}

Steps~A--D fix the configuration: the rubric and weights, the prompt, the
context, and the generation model. Steps~E--F set up evaluation, Step~G
runs the task and scores the results, and Step~H decides whether to
release. Model selection at Step~D is itself an evaluation pass, using
the judge of Step~E and the scoring procedure of Step~G applied to each
candidate in turn.

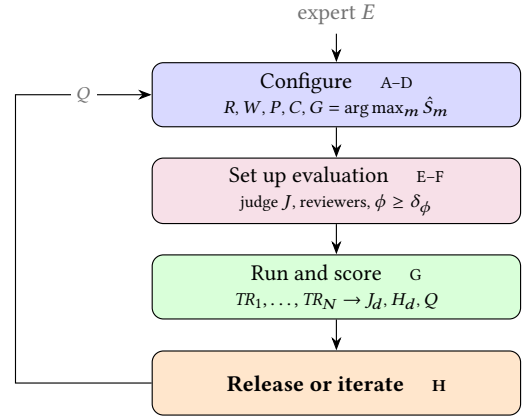
\begin{figure}[h!]
\centering
\resizebox{0.8\columnwidth}{!}{%
\begin{tikzpicture}[
    box/.style={draw, rounded corners, minimum width=4.6cm,
                minimum height=0.8cm, align=center, font=\small},
    final/.style={draw, rounded corners, fill=orange!20,
                  minimum width=4.6cm, minimum height=0.8cm,
                  align=center, font=\small\bfseries},
    arr/.style={-{Stealth}}
]
\node[font=\small, text=black!55] (E) at (0, 5.2) {expert $E$};
\node[box, fill=blue!15]  (cfg) at (0, 4.2)
  {Configure \quad {\scriptsize A--D}\\[-1pt]{\scriptsize $R$,\ $W$,\ $P$,\ $C$,\ $G = \arg\max_m \hat{S}_m$}};
\node[box, fill=purple!12] (ins) at (0, 3.0)
  {Set up evaluation \quad {\scriptsize E--F}\\[-1pt]{\scriptsize judge $J$,\ reviewers,\ $\phi \geq \delta_\phi$}};
\node[box, fill=green!15] (gen) at (0, 1.8)
  {Run and score \quad {\scriptsize G}\\[-1pt]{\scriptsize $\mathit{TR}_1,\ldots,\mathit{TR}_N$\ $\rightarrow$\ $J_d$,\ $H_d$,\ $Q$}};
\node[final]              (dec) at (0, 0.6)
  {Release or iterate \quad {\scriptsize H}};
\draw[arr] (E)   -- (cfg);
\draw[arr] (cfg) -- (ins);
\draw[arr] (ins) -- (gen);
\draw[arr] (gen) -- (dec);
\draw[-{Stealth}]
  (dec.west) -- ++(-1.7, 0) -- ++(0, 3.6) -- (cfg.west)
  node[pos=0.5, fill=white, font=\scriptsize, text=black!55, inner sep=2pt]
    {$Q$};
\end{tikzpicture}
}%
\caption{Framework pipeline. The expert configures the task, rubric, and
generation model, evaluation is set up, the task is run and scored, and a
release decision is made. The quality vector $Q$ returns to the
configuration stage.}
\label{fig:framework}
\end{figure}

%% =====================================================================
\subsection{Step A. Rubric and Weights}

The rubric defines the criteria against which each task result is scored,
translating output quality into discrete, scoreable dimensions. At the
first iteration, $\tau = 0$, it is written by domain experts; in
subsequent iterations it is revised using the previous quality vector:

\begin{gather}
    R^{(0)} = R(E), \qquad R^{(\tau)} = R\!\left(E,\, Q^{(\tau-1)}\right) \\
    W^{(0)} = W(E), \qquad W^{(\tau)} = W\!\left(E,\, Q^{(\tau-1)}\right)
\end{gather}

\noindent Weights are revised on the same schedule and for the same
reason. A dimension that scored poorly at $\tau - 1$ may be overweighted
so that the next iteration's model selection and prompt revision are
pulled toward it. Weight changes are recorded on the card with their
rationale, so that a selection outcome can be traced to a stated priority
grounded in prior results rather than to an unexplained retuning.
Dimensions themselves may be added, removed, or respecified as
understanding of the task deepens \cite{llm_rubric_design_1,
llm_rubric_design_2, llm_rubric_design_3}. Because a rubric change alters
the measurement instrument, the deploying team should note which
dimensions carried over unchanged; scores are comparable across
iterations only on those.

%% NOTE: Restored the carry-over sentence, in the team's voice. Without
%% it, Delta J_d and Delta H_d in Step G are undefined wherever the
%% rubric changed, which it did between your tau = 0 and tau = 1.

%% =====================================================================
\subsection{Step B. Prompt}

The prompt is the primary instruction to $G$, specifying what the task
result should contain, how it should be structured, and what constraints
it must satisfy:

\begin{equation}
    P^{(0)} = P(E),
    \qquad
    P^{(\tau)} = P\!\left(E,\, Q^{(\tau-1)}\right)
\end{equation}

\noindent Later iterations use $Q^{(\tau-1)}$ to direct targeted
revisions at the dimensions that scored lowest.

%% =====================================================================
\subsection{Step C. Context}

The context supplies $G$ with domain knowledge beyond the prompt:
reference materials, worked examples, data constraints, or precedents. It
is optional, and is revised on the same schedule:

\begin{equation}
    C^{(\tau)} = C\!\left(E,\, Q^{(\tau-1)}\right),
    \qquad C \in \{\cdot, \emptyset\}
\end{equation}

\noindent One option during model selection (Step~D) is to withhold
context, setting $C = \emptyset$ across all candidates, and measure
performance using zero-shot prompts. Whichever the deploying team
chooses, the card records it, since the two conditions answer different
questions: withheld context compares the models, supplied context
compares the deployments.

%% =====================================================================
\subsection{Step D. Generation Model Selection}

The generation model is chosen from a candidate set $\mathcal{M}$ by
running the evaluation of Steps~E--G for each candidate, with $P$, $R$,
$W$ held fixed and $C$ handled as described in Step~C. This gives, for
each candidate $m$, a quality score $Q_d(m)$ per dimension plus mean cost
$c_m$ and latency $t_m$. Each dimension is min-max normalized across
candidates before being combined.\footnote{$\tilde{x}_m = \left(x_m -
\min_{m'} x_{m'}\right) / \left(\max_{m'} x_{m'} - \min_{m'}
x_{m'}\right)$, applied across all $m \in \mathcal{M}$.} The composite
score and selection rule mirror Equation~\eqref{eq:objective}:

\begin{gather}
    \hat{S}_m = \sum_{d=1}^{D} w_d\, \tilde{Q}_d(m)
    - w_c\, \tilde{c}_m - w_t\, \tilde{t}_m \\
    G = \arg\max_{m \in \mathcal{M}} \hat{S}_m
\end{gather}

The weights make domain priorities explicit and auditable. Raising $w_d$
on a dimension that underperformed at $\tau - 1$ lets a selection
decision be justified by pointing to a stated priority grounded in prior
results rather than to an opaque aggregate.

Where the deploying team treats cost or latency as a hard limit rather
than something to trade off, the weighted form can be replaced by
maximizing $\sum_d w_d Q_d(m)$ subject to $c_m \leq B_c$ and
$t_m \leq B_t$, with the budgets recorded on the card.

\subsection{Step E. LLM Judge}

A second model is designated as the judge $J$. Where $G$ produces a task
result, $J$ scores it against the rubric and weights. To limit
self-preference bias, the tendency of a model to favor outputs from its
own family \cite{llm_judge_self_bias_1, llm_judge_self_bias_2,
llm_judge_self_bias_3}, the judge should satisfy:

\begin{equation}\label{eq:judge}
    \mathcal{F}(J) \neq \mathcal{F}(G),
    \qquad
    \rho(J) \geq \rho(G)
\end{equation}

\noindent The first condition puts the judge in a different family from
the generator. The second keeps the judge from being less equipped to
catch errors than the generator was to produce them. Neither is
enforceable, and a deploying team may have good reason to violate either,
most often cost or availability.

Judge scores are stochastic, so the judge is run $\ell$ times per task
result and the spread across runs is reported alongside $J_d$. A
cache-busting protocol ensures repeated runs are not served from cache.

\subsection{Step F. Human Review and Rubric Reliability}

Human reviewers may score task results in addition to or in place of $J$.

With more than one reviewer, inter-rater reliability $\phi$ determines
whether their scores are trustworthy enough to use. The choice of metric
follows from the scale set in Step~A rather than being fixed by the
framework. For a dimension scored on defined ordinal levels, ordinal
Krippendorff's $\alpha$ is used since it tolerates a varying number of
reviewers per item. Reliability is computed per dimension and overall,
and the deploying team may set an optional threshold $\delta_\phi$ gating
entry into $Q$:

\begin{equation}
    H_d \text{ is admitted to } \mathrm{agg} \iff \phi \geq \delta_\phi
\end{equation}

\noindent Below the threshold, reviewers might be applying the rubric
inconsistently. A development team may view this as evidence of rubric
ambiguity rather than genuine expert disagreement, and revise the rubric
before admitting human scores; per-dimension $\phi$ identifies which
dimensions need clarification. The opposite reading is also available and
may be the right one: low agreement can mean the dimension is genuinely
contested among experts, in which case reporting the dispersion is more
faithful than revising it away. Which reading applies is a domain
judgment, and the card records the interpretation taken alongside the
figure itself.

When $\phi$ is undefined for a dimension because every score is
identical, the development team might check for a ceiling effect. A
constant score may reflect a fixed evaluator response rather than true
agreement, and such a dimension carries no signal, so it is flagged for
revision rather than counted as perfect reliability.

\subsection{Step G. Run and Score}

At each iteration $G$ runs the task on all $N$ inputs:

\begin{equation}
    \mathit{TR}_n = G\!\left(P,\, C,\, R,\, Q^{(\tau-1)}\right),
    \qquad n = 1,\ldots,N
\end{equation}

\noindent Each result is scored by the judge, by human reviewers, or
both, and the scores are combined as described in
Section~\ref{sec:setup} to give $J_d$, $H_d$, and $Q$. Four diagnostics
are reported per dimension and carried onto the card:

\begin{itemize}[leftmargin=*, itemsep=2pt, topsep=3pt]
  \item $J_d$ and $H_d$, with their standard deviations across judge runs
  and across reviewers respectively;
  \item the judge--human gap $J_d - H_d$;
  \item the change from the previous iteration, $\Delta J_d$ and
  $\Delta H_d$, defined only on dimensions carried unchanged from
  $\tau - 1$ (Step~A).
\end{itemize}

\noindent The gap is the quantity that directs expert effort. A dimension
where the judge tracks human scores closely can be delegated to $J$ in
later iterations; a dimension with a large gap needs continued human
review. Judge spread separates an unstable judge from a genuinely
difficult dimension.

%% =====================================================================
\subsection{Step H. Release or Iterate}

Release requires that quality clear a threshold on every dimension, not
just on average. Writing $\bar{J} = \frac{1}{D}\sum_d J_d$ and
$\bar{H} = \frac{1}{D}\sum_d H_d$ for the overall scores:

\begin{gather}
\text{release} \iff
\bar{J} \geq \delta_J \;\text{ and }\; \bar{H} \geq \delta_H \\
\text{and} \quad
J_d \geq \delta_{J,d} \;\text{ and }\; H_d \geq \delta_{H,d}
\;\;\; \text{for all } d \notag
\end{gather}

\noindent The deploying team may choose to apply the human conditions
only when $H_d$ passes the Step~F gate. Otherwise the pipeline advances
to $\tau + 1$ and $Q$ informs revisions to $P$, $C$, $R$, and $W$.

The thresholds themselves are the team's to set, and setting them is the
point at which an abstract commitment to quality becomes a number
somebody has to defend. What the framework fixes is the shape: floors
apply per dimension, not only to the average. A threshold on the average
alone lets a system release while failing on one dimension. Floors may differ across dimensions, and a dimension needing no
floor is assigned $0$ and recorded as such.

\paragraph{Termination.}
The loop is not guaranteed to reach a releasable configuration. A
deploying team may stop iterating for reasons the framework does not
model, such as an exhausted budget or a judgment that the remaining gap
will not close with further prompt and rubric revision. Stopping without
release is a legitimate outcome rather than a failure of the framework,
and the final card documents it: the configuration reached, which
dimensions failed their floors, and the team's diagnosis of why. Recorded
that way, the decision can be revisited when a more capable candidate
appears, without the reasoning having to be reconstructed from memory.

\newcommand{\cho}{\textsc{choice}}
\newcommand{\der}{\textsc{derived}}
\newcommand{\req}{\textsc{record}}
\newcommand{\opt}{\textsc{optional}}

\subsection{The MERCI Card}\label{sec:card}

Each iteration closes by issuing a MERCI ((Model-Evaluation-Rubric-Continuous-Improvement) card: a single versioned record
of the configuration that was run, the scores it produced, and the
decision that followed. The card is the framework's enforcement
mechanism. Section~\ref{sec:framework} leaves the substance of every
configurable choice to the deploying team, so what remains to be required
is that the choices be legible afterwards, to a colleague inheriting the
system, to a reviewer assessing a claim made on its output, or to the
team itself deciding a year later whether a new model release is worth
adopting.

Table~\ref{tab:merci} gives the specification. Each field is labelled by
what kind of thing it is. A \textsc{choice} is a decision the team makes,
and recording it means recording both the value and the reasoning behind
it; these are the fields of Table~\ref{tab:decisions}. A \textsc{derived}
field is computed from the run and is not open to interpretation at
recording time. A \textsc{record} is a factual matter such as a date, a
model identifier, or a prior card reference. \textsc{Optional} fields
improve the card's usefulness without being necessary to interpret the
iteration.

The card is written per iteration rather than per release, which means
iterations that end without release generate cards too. A sequence of
cards is therefore a history of what was tried, not a record of what
succeeded: a configuration abandoned at $\tau = 2$
is worth knowing about when a similar one is proposed at $\tau = 5$.

\begin{table*}[h]
\centering
\footnotesize
\caption{MERCI Card specification. One card per iteration $\tau$.
\textsc{choice} fields are decisions the deploying team makes: record the
value and the reasoning. \textsc{derived} fields are computed from the
run. \textsc{record} fields are factual. \textsc{optional} fields improve
the card without being necessary to interpret the iteration.}
\label{tab:merci}
\setlength{\tabcolsep}{5pt}
\renewcommand{\arraystretch}{1.05}
\begin{tabular}{@{}l p{10.4cm} >{\raggedright\arraybackslash}p{1.5cm}@{}}
\toprule
\textbf{Section} & \textbf{Field} & \textbf{Kind} \\
\midrule

\multirow{3}{*}{\textbf{Identity}}
  & System name, domain, iteration $\tau$                                    & \req \\
  & Card version, date issued, prior card reference ($\tau-1$)              & \req \\
  & Deploying team and accountable contact                                  & \req \\
\midrule

\multirow{4}{*}{\textbf{Task}}
  & $\mathcal{T}$: task description and what a task result is
    (e.g., answer, report, summary, drafted document)                       & \req \\
  & $N$: number of inputs the task was run on; how they were selected       & \req \\
  & Scoring granularity per dimension: continuous, discrete, or
    categorical, with the mapping onto $[0,1]$                              & \cho \\
  & $\mathrm{agg}$: how $J_d$ and $H_d$ combine into $Q_d$; $\lambda$
    if weighted                                                             & \cho \\
\midrule

\multirow{4}{*}{\textbf{Rubric and Weights}}
  & $R = \{d_1,\ldots,d_D\}$: dimensions with descriptions                  & \cho \\
  & $W = \{w_1,\ldots,w_D\}$: dimension weights                             & \cho \\
  & $E$: who defined $R$, and their domain expertise                        & \req \\
  & Dimensions added, removed, or respecified since $\tau-1$, and which
    carried over unchanged                                                  & \req \\
\midrule

\multirow{2}{*}{\textbf{Prompt}}
  & $P$: prompt text or a reference to the versioned artifact               & \cho \\
  & Revisions since $\tau-1$, and which dimensions of $Q^{(\tau-1)}$
    motivated each                                                          & \req \\
\midrule

\multirow{2}{*}{\textbf{Context}}
  & $C$: what was supplied, or $\emptyset$; provenance of each source       & \cho \\
  & Whether $C$ was withheld during model selection, and why                & \cho \\
\midrule

\multirow{7}{*}{\textbf{Model Selection}}
  & $\mathcal{M}$: candidate set in full, listed individually               & \cho \\
  & Mapping from each family's native settings onto
    $\rho(m) \in \{0,\ldots,4\}$                                            & \cho \\
  & $w_d$, $w_c$, $w_t$, or budgets $B_c$, $B_t$ under the
    constrained form                                                        & \cho \\
  & Aggregation used for selection, where it differs from the deployed
    $\mathrm{agg}$ (e.g., $Q_d(m) = J_d(m)$), and the share of the
    sweep that received human review                                        & \cho \\
  & $Q_d(m)$, $c_m$, $t_m$ for every $m \in \mathcal{M}$                    & \der \\
  & $\hat{S}_m$ for every $m \in \mathcal{M}$, post-normalization; metrics
    on which all candidates tied and were dropped                           & \der \\
  & $G = \arg\max_m \hat{S}_m$, with the margin over the runner-up          & \der \\
\midrule

\multirow{5}{*}{\textbf{LLM Judge}}
  & $J$: judge model, $\mathcal{F}(J)$, $\rho(J)$                           & \cho \\
  & $\ell$: judge runs per task result; cache-busting protocol              & \cho \\
  & $\mathcal{F}(J) \neq \mathcal{F}(G)$: satisfied or violated; if
    violated, the implication for the scores                                & \req \\
  & $\rho(J) \geq \rho(G)$: satisfied or violated, with implication         & \req \\
  & $J_d$ per dimension, with standard deviation across the $\ell$ runs     & \der \\
\midrule

\multirow{7}{*}{\textbf{Human Review}}
  & $V$: number of reviewers; expertise and recruitment method              & \cho \\
  & Scoring protocol: independence, blinding, order effects                 & \opt \\
  & $\phi$: reliability metric chosen, and why it fits the scale            & \cho \\
  & $\delta_\phi$: threshold, if the team set one                           & \cho \\
  & $\phi$ per dimension and overall; dimensions where $\phi$ is
    undefined for want of variance                                          & \der \\
  & $H_d$ per dimension, with inter-reviewer standard deviation             & \der \\
  & Where $\phi < \delta_\phi$: whether read as rubric ambiguity or as a
    genuinely contested dimension, and what followed                        & \cho \\
\midrule

\multirow{4}{*}{\textbf{Run and Score}}
  & $\mathit{TR}_1,\ldots,\mathit{TR}_N$: task results, or a reference
    to where they are stored                                                & \req \\
  & $Q = [\,Q_1,\ldots,Q_D\,]$: quality vector                              & \der \\
  & $J_d - H_d$ per dimension: judge--human gap                             & \der \\
  & $\Delta J_d$, $\Delta H_d$: change from $\tau-1$, on carried-over
    dimensions only                                                         & \der \\
\midrule

\multirow{5}{*}{\textbf{Release Decision}}
  & $\delta_J$, $\delta_H$: overall thresholds                              & \cho \\
  & $\delta_{J,d}$, $\delta_{H,d}$: per-dimension floors, $0$ where none
    applies; whether the human conditions were applied at all               & \cho \\
  & Decision: release, or advance to $\tau+1$; dimensions that failed
    their floors                                                            & \der \\
  & Rationale for the decision                                              & \req \\
  & Planned changes to $R$, $W$, $P$, $C$, $G$ for $\tau+1$;
    \texttt{null} on release or on stopping without release                 & \opt \\
\midrule

\multirow{5}{*}{\textbf{Limitations}}
  & Release scope, and the human oversight required before a task
    result is relied on                                                     & \req \\
  & Judge constraint violations carried from the judge section, and their
    implications read together                                              & \req \\
  & Dimensions showing ceiling effects, distinguished from dimensions
    with genuine agreement                                                  & \req \\
  & Data constraints and known exclusions affecting task result quality     & \req \\
  & Where the team stopped iterating without release: the diagnosis of
    why the gap did not close                                               & \opt \\
\bottomrule
\end{tabular}
\end{table*}

\section{Example Use Cases}\label{sec:application}

\subsection{Redo.io}\label{sec:redo}

\paragraph{Background.}
The California Racial Justice Act allows a person to challenge a
conviction or sentence by showing racial disparity in how similar cases
were charged or sentenced. At the prima facie stage the showing is
statistical, which means a defender must assemble sentencing data,
select appropriate tests, and present the results in a form a court will
credit. Public defender offices carry heavy caseloads and rarely have
statistical staff, so the effort a single filing requires is often the
binding constraint on whether a claim is brought at all
\cite{ball2025, bazelon2025, kuluk2024_impact}.

\paragraph{The system.}
Redo.io is a nonprofit platform that produces statistical evidence
reports for these filings. Disparity statistics are computed upstream by
conventional code: odds ratios, relative risks, chi-square tests, with
their confidence intervals, $p$-values, and sample sizes. The language
model's role is interpretive rather than analytical. It receives the
computed figures and drafts a narrative that states what they show,
where they disagree, and what the underlying data cannot support. Every
report requires attorney review before it is filed.

\paragraph{Task and task result.}
The task $\mathcal{T}$ is to synthesize a set of pre-computed statistics
into an evidence narrative. A task result $\mathit{TR}$ is one such
report, covering one sentencing county, sentencing outcome, and ethnicity
pair, with cohort controls on prior record. Reports are the unit of
evaluation throughout.

\subsection{Instantiating the Framework}\label{sec:instantiation}

We report the second iteration, $\tau = 1$. A first iteration established
the rubric from expert input alone, without model selection or human
review, and produced the quality vector $Q^{(0)}$ that drove the
revisions below. Appendices~B, C, and J give the prompt, rubric, and
reliability analysis; source materials are at
\url{https://github.com/redoio/ai_evals}.

\paragraph{Rubric and weights.}
The rubric has $D = 9$ dimensions (Table~\ref{tab:rubric}): six on
correct statistical contextualization, two on stating what the data
cannot support, and one, Attribution, on whether disparities are
attributed to systemic factors rather than to the propensity of any
racial group \cite{propublica_machinebias, rudin2020_secrecy,
rudin2019_blackbox, miron2020}. All nine carried over unchanged from
$\tau = 0$, so scores are comparable across iterations. Weights did not:
cross-method comparison was weakest at $\tau = 0$, so we set
$w_{\text{Cmp}} = 5$, with $w_d = 1$ otherwise and $w_c = w_t = 1$.

\begin{table}[t]
\centering
\footnotesize
\caption{Examples of evaluation rubric dimensions}
\label{tab:rubric}
\setlength{\tabcolsep}{4pt}
\renewcommand{\arraystretch}{1.1}
\begin{tabular}{@{}l p{5.6cm}@{}}
\toprule
\textbf{Dimension} & \textbf{Tests whether the report} \\
\midrule
OR CI\textsubscript{score}    & contextualizes odds ratios with CI width
                                and whether the interval spans 1 \\
RR CI\textsubscript{score}    & does the same for relative risk \\
OR Smp\textsubscript{score}   & contextualizes odds ratios with sample
                                size \\
RR Smp\textsubscript{score}   & does the same for relative risk \\
$\chi^2$ Smp\textsubscript{score} & does the same for chi-square \\
$p$\textsubscript{score}      & contextualizes $p$-values against the
                                significance threshold \\
Cmp\textsubscript{score}      & compares the statistical methods and reconciles methods that diverge \\
Lim\textsubscript{score}      & states dataset limitations and their
                                effect on the findings \\
Att\textsubscript{score}      & attributes disparities to systemic
                                factors, not inherent criminality \\
\bottomrule
\end{tabular}
\end{table}

\paragraph{Prompt and context.}
The prompt directs $G$ to interpret the supplied figures across the three
statistical methods, surface disagreement between them, and
attribute any disparity to systemic factors; it states explicitly that
$G$ must perform no computation. The context supplies excerpts from prior
legal filings and documentation of known gaps in the underlying sentencing dataset.

\paragraph{Model selection.}
We evaluated $|\mathcal{M}| = 11$ configurations over $N = 18$ reports at
$\ell = 15$ judge runs each, judge scores only, context withheld.
Table~\ref{tab:model_performance} declares the
mapping from each family's native setting onto
$\rho \in \{0,\ldots,4\}$, so a reader who would map them differently can
see what was assumed. The selected configuration is Opus at
$\rho(G) = 3$, $\hat{S}_G = 11.94$.

\paragraph{Judge and human review.}
The judge is GPT at $\rho(J) = 3$, satisfying both conditions. It was
fixed across the sweep while $\mathcal{M}$ spans two families, so three
of eleven candidates were scored within family and eight were not. Each
report is scored $\ell = 15$ times with a cache-busting
protocol.\footnote{Each request carries a fresh UUID for
\texttt{user} and \texttt{prompt\_cache\_key}, routing repeated requests
to different cache buckets.} Three domain experts then scored the
reports independently. We set no $\delta_\phi$ in advance.

\begin{table}[t]
\centering
\small
\caption{Model selection sweep: $|\mathcal{M}| = 11$ configurations over
$N = 18$ reports and $\ell = 15$ judge runs. Native settings appear with
the normalized reasoning level $\rho$ the team declared for them. $J_d$
is the mean judge score across the nine dimensions; $\hat{S}_m$ is the
composite including cost and latency ($w_{\text{Cmp}} = 5$, $w_d = 1$
otherwise, $w_c = w_t = 1$). Sorted by $J_d$.
\colorbox{green!20}{Highlighted row} is the selected $G$.}
\label{tab:model_performance}
\setlength{\tabcolsep}{3pt}
\footnotesize
\begin{tabular}{@{}l l c c c r r r r@{}}
\toprule
\multicolumn{3}{c}{Generation model} & \multicolumn{2}{c}{Judge}
& & & & \\
\cmidrule(lr){1-3}\cmidrule(lr){4-5}
$\mathcal{F}(G)$ & native & $\rho$ & $\mathcal{F}(J)$ & $\rho$
& $J_d$ & $\hat{S}_m$ & $t$ (s) & $c$ (\$) \\
\midrule
Opus   & none          & 0 & GPT  & 3 & 0.92 & 11.60 & 40  & 0.06 \\
Opus   & adapt / med   & 2 & GPT  & 3 & 0.92 & 10.92 & 53  & 0.08 \\
\rowcolor{green!20}
Opus   & adapt / high  & 3 & GPT  & 3 & 0.92 & 11.94 & 55  & 0.08 \\
Opus   & adapt / max   & 4 & GPT  & 3 & 0.91 & 10.31 & 96  & 0.16 \\
Sonnet & adapt / high  & 3 & GPT  & 3 & 0.90 &  9.91 & 180 & 0.21 \\
Sonnet & adapt / med   & 2 & GPT  & 3 & 0.90 &  9.10 & 166 & 0.19 \\
Sonnet & adapt / max   & 4 & GPT  & 3 & 0.88 &  8.48 & 274 & 0.34 \\
Sonnet & none          & 0 & GPT  & 3 & 0.85 &  8.19 & 38  & 0.04 \\
\midrule
GPT    & high          & 3 & Opus & 4 & 0.82 &  8.12 & 96  & 0.09 \\
GPT    & med           & 2 & Opus & 3 & 0.81 &  8.63 & 54  & 0.05 \\
GPT    & none          & 0 & Opus & 3 & 0.67 &  1.52 & 25  & 0.02 \\
\bottomrule
\end{tabular}
\end{table}

\begin{table}[t]
\centering
\footnotesize
\caption{Scores per rubric dimension at $\tau = 1$ ($N = 30$,
$\ell = 15$, $V = 3$), with the change from the prior iteration. Judge sd
is the mean standard deviation across the $\ell$ runs per report; human
sd is the inter-reviewer standard deviation.}
\label{tab:dim_scores}
\setlength{\tabcolsep}{4pt}
\begin{tabular}{@{}l rr rr c rr@{}}
\toprule
& \multicolumn{2}{c}{Judge} & \multicolumn{2}{c}{Human}
& & \multicolumn{2}{c}{$\Delta$ from $\tau = 0$} \\
\cmidrule(lr){2-3}\cmidrule(lr){4-5}\cmidrule(lr){7-8}
\textbf{Dimension} & $J_d$ & sd & $H_d$ & sd
& & $\Delta J_d$ & $\Delta H_d$ \\
\midrule
OR CI\textsubscript{score}        & 0.94 & 0.03 & 0.89 & 0.16 && $+$0.15 & $+$0.09 \\
OR Smp\textsubscript{score}       & 1.00 & 0.00 & 0.80 & 0.13 && $+$0.51 & $+$0.19 \\
RR CI\textsubscript{score}        & 0.86 & 0.04 & 0.87 & 0.18 && $+$0.10 & $+$0.07 \\
RR Smp\textsubscript{score}       & 1.00 & 0.00 & 0.80 & 0.13 && $+$0.51 & $+$0.19 \\
Cmp\textsubscript{score}          & 0.74 & 0.01 & 0.76 & 0.35 && $+$0.05 & $-$0.02 \\
$\chi^2$ Smp\textsubscript{score} & 1.00 & 0.02 & 0.80 & 0.14 && $+$0.50 & $+$0.21 \\
$p$\textsubscript{score}          & 1.00 & 0.00 & 0.99 & 0.01 && $+$0.11 & $+$0.15 \\
Lim\textsubscript{score}          & 0.75 & 0.00 & 0.96 & 0.07 && \pz0.00 & $+$0.02 \\
Att\textsubscript{score}          & 0.96 & 0.02 & 1.00 & 0.00 && $-$0.04 & \pz0.00 \\
\midrule
\textbf{Overall}                  & 0.92 & 0.01 & 0.88 & 0.13 && $+$0.21 & $+$0.10 \\
\bottomrule
\end{tabular}
\end{table}

\subsection{The Resulting MERCI Card}\label{sec:filled_card}

Table~\ref{tab:filled_card} is the card for this iteration: not a claim
that the system works, but a statement of what was configured, what it
scored, what was wrong with the measurement, and what followed. Two
entries do work the score tables cannot. The fivefold weight sits beside
the modest gain it produced, making the intervention assessable rather
than merely reported. And the reliability figure sits beside the reading
taken from it, so a later reader can disagree with the reading without
re-deriving the number.

The card records no release. We set no per-dimension floors in advance,
which is itself a finding: the framework asks for them before the scores
arrive, precisely so a threshold cannot be chosen to fit a result. Under
a uniform $0.80$ floor this configuration would have failed on
cross-method comparison and on limitations. Reporting that is more useful
than reporting the $0.92$ average alone, and it is what the
per-dimension rule exists to surface.

\begin{table*}[t]
\centering
\footnotesize
\caption{MERCI Card for $\tau = 1$.}
\label{tab:filled_card}
\setlength{\tabcolsep}{5pt}
\renewcommand{\arraystretch}{1.1}
\begin{tabular}{@{}l l p{10.2cm}@{}}
\toprule
\textbf{Section} & \textbf{Field} & \textbf{Entry} \\
\midrule

\multirow{2}{*}{\textbf{Identity}}
  & Prior card & $\tau = 0$ \\
  & $N$ inputs & 30 reports; 18 in the selection sweep \\
\midrule

\multirow{2}{*}{\textbf{Task}}
  & Granularity & any value in the range $[0,1]$ \\
  & $\mathrm{agg}$ & $Q_d = \tfrac{1}{2}(J_d + H_d)$ for deployment evaluation, i.e. both humans and LLM judges are used;
  $Q_d = J_d$ for model selection, i.e. no humans are used \\
\midrule

\multirow{3}{*}{\textbf{Rubric}}
  & $R$ & $R(E, Q^{(0)})$, $D = 9$ \\
  & $W$ & $w_{\text{Cmp}} = 5$, $w_d = 1$ otherwise, $w_c = w_t = 1$ \\
  & Rationale & Cmp was weakest at $\tau = 0$ ($J_d = 0.69$) so additional weights applied\\
\midrule

\multirow{2}{*}{\textbf{Prompt, Context}}
  & $P$, $C$ & $P(E, Q^{(0)})$; $C(E, Q^{(0)})$, context on data limitations and comparison for statistical methods expanded since these were
    lowest-scoring dimensions in $\tau = 0$\\
  & In selection & context withheld, $C = \emptyset$ \\
\midrule

\multirow{4}{*}{\textbf{Model Selection}}
  & $\mathcal{M}$ & 11 configurations, listed in Table~\ref{tab:model_performance} \\
  & $\rho$ mapping & declared in Table~\ref{tab:model_performance} \\
  & $G$ & Opus, native \textit{adapt / high}, $\rho(G) = 3$ \\
  & $\hat{S}_G$ & $11.94$ out of $14$, the best possible score \\
\midrule

\multirow{3}{*}{\textbf{LLM Judge}}
  & $J$, $\ell$ & GPT, $\rho(J) = 3$; $\ell = 15$; cache-busting per run \\
  & Constraints & $\mathcal{F}(J) \neq \mathcal{F}(G)$ satisfied;
    $\rho(J) \geq \rho(G)$ satisfied ($3 \geq 3$)  \\
\midrule

\multirow{3}{*}{\textbf{Human Review}}
  & $V$, $\phi$ & 3 expert reviewers; ordinal Krippendorff's $\alpha$ \\
  & $\delta_\phi$ & \textbf{not set in advance} \\
  & Result and reading & $\alpha = 0.08$ overall, negative for 2 of 3
    pairs; read as rubric ambiguity, calibration required before
    $\tau = 2$ \\
\midrule

\multirow{3}{*}{\textbf{Run and Score}}
  & $\bar{J}$, $\bar{H}$ & $0.92$; $0.88$ ($\Delta$ from $\tau = 0$: $+0.21$, $+0.10$ respectively) \\
  & Gap $J_d - H_d$ & $+0.04$ overall; $-0.21$ on Limitations,
    $-0.20$ on the sample size dimensions \\
  & Weakest $d$ & Cmp ($0.74$), Limitations ($0.75$) \\
\midrule

\multirow{3}{*}{\textbf{Release}}
  & Floors $\delta_{\cdot,d}$ & \textbf{not set in advance} \\
  & Decision & iterate \\
  & Rationale & Cmp and Limitations dimensions perform poorly; human reliability
    unusable due to poor agreement \\
\midrule

\multirow{3}{*}{\textbf{Limitations}}
  & Ceiling dimensions & 4 of 9 show zero judge variance across 450
    evaluations \\
  & Measurement & human scores not admitted as a reliable signal this
    iteration \\
  & Oversight required & attorney review before any report is filed \\
\bottomrule
\end{tabular}
\end{table*}

\subsection{Framework Scope and Transferability}

The framework presented here is domain-agnostic. The core requirements for adoption are a rubric definable and refinable by domain experts, a generation task for which LLM judges can approximate human judgment, and a cost and latency budget that makes exhaustive human evaluation impractical at every iteration. The framework is particularly well-suited to under-resourced legal or related settings where the weighted composite scores makes cost and time selection criteria rather than afterthoughts.

Our findings point to two transferable insights. First, the separation of roles between LLM judges as a scalable, reproducible evaluation instrument and human evaluators as a signal allows practitioners to concentrate human effort on dimensions where it has the highest marginal value---specifically those with high inter-rater spread or large human-machine divergence. Second, the weight vector provides an explicit, auditable mechanism for encoding domain priorities: model selection decisions can be justified by pointing to weight choices grounded in prior evaluation outcomes rather than opaque aggregate scores. In the Redo.io case study, elevating $w_{\text{Cmp}}$ at $\tau = 1$ directly reflected a practitioner judgment derived from $\tau = 0$, demonstrating that the MERCI Cards framework can support evidence-based configuration decisions rather than arbitrary tuning.

More broadly, the pace of general-purpose model releases means that any deployed pipeline faces continuous pressure to re-evaluate whether the current generation task remains optimal. The iterative structure of the framework anchored by the feedback loop provides a stable procedure within which new models can be tested and integrated without redesigning the evaluation pipeline from scratch. This property is especially valuable in high-stakes settings, where the frontier of model capabilities can translate directly into improved outcomes for those the system is designed to serve but resources for end-to-end evaluation are severely constrained.

\section{Limitations of MERCI Cards}

While MERCI Cards provide a structured artifact for documenting iterative LLM deployment, three limitations warrant acknowledgment. First, the card is designed for developers and researchers familiar with the underlying framework: many fields in the card require fluency with the mathematical notation and pipeline design to interpret correctly. Practitioners without a technical background (including the attorneys and advocates who are the intended beneficiaries of systems like Redo.io) cannot readily read a MERCI Card to assess whether a system is ready for use. A practitioner-facing summary layer, analogous to the "intended use" and "out-of-scope use" sections of model cards, would need to accompany the card for it to serve a broader audience \cite{model_cards_ml}.

Second, the card records what happened at each iteration but cannot
establish why. A low inter-rater reliability score may
reflect ambiguous rubric language, insufficient reviewer training, genuine domain disagreement among experts, or reviewers with varying levels of expertise---and the card does not distinguish between these explanations. Similarly, a gain in $\Delta L$ across iterations may be attributable to the generation model upgrade, the addition of context $C^{(\tau)}$, or rubric revision, or all three simultaneously. Without a controlled study, the card captures correlation across iterations but not causal attribution.

Third, the card structure reflects the priorities of the development team at the time of its design, which may shift as the system matures. Rubric dimensions that are critical early in deployment may become less
informative as performance ceilings are reached, while new dimensions
reflecting evolving legal standards or practitioner needs may need to be introduced. The card must therefore be treated as a living document,
versioned alongside the pipeline it documents, rather than a fixed
specification applied uniformly across iterations.

\section{Conclusion}

We present a mathematical framework for iterative LLM integration in high-stakes, resource-constrained domains, formalizing model selection, rubric design, and evaluation feedback as a principled loop in which outcomes drive configuration decisions at each iteration. Applied to RJA evidence synthesis on the \href{https://getredo.io}{Redo.io} platform, the framework yields substantial performance improvements across iterations.

\bibliographystyle{ACM-Reference-Format}
\bibliography{sample-base}

\appendix

\section{GenAI Use Disclosure}

Authors leveraged Claude and GPT to assist with generating LaTeX code for tables, and grammar and cohesion checks.

\section{Example of AI-Powered Evidence Report}

Example of a report generated using \url{https://app.getredo.io/bias_analysis}:
\begin{quote}
\textbf{Executive Summary}
The sentencing outcome analyzed is sentence type "DSL" and the comparison is between White and Hispanic individuals. There is evidence of a sentencing disparity that disproportionately results in Hispanic individuals receiving DSL; the disparity is moderate in magnitude and warrants further scrutiny of systemic sentencing processes.

\textbf{Findings}\\
\textit{What statistical methods were used?}\\
The odds ratio was used to compare the odds that White individuals receive DSL relative to Hispanic individuals; it expresses the ratio of odds of the outcome across the two groups. The relative risk was used to compare the probability (risk) of receiving DSL for White individuals relative to Hispanic individuals; it provides an intuitive measure of how much more or less likely the outcome is in one group versus the other. The chi-square test of independence was used to evaluate whether an association exists between ethnic group and sentencing outcome, testing whether the observed distribution of cases across groups departs from what would be expected under independence.

\textit{What meaning and trends did we find?}\\
The odds ratio of 0.521 with 95\% confidence interval (0.364, 0.745) indicates that the odds of White individuals receiving DSL are lower than the odds for Hispanic individuals, and the confidence interval excludes 1.0, indicating statistical significance at the conventional 0.05 level. The relative risk of 0.69 with 95\% confidence interval (0.566, 0.857) indicates that the risk of receiving DSL is lower for White individuals compared to Hispanic individuals, and the confidence interval excludes 1.0, indicating statistical significance. The chi-square test produced a statistic of 12.26 with p-value 0.0005, providing strong evidence of an association between ethnic group and the DSL sentencing outcome that is unlikely to be due to chance. Taken together, the three methods are concordant: they show a statistically significant disparity in DSL sentencing that disfavors Hispanic individuals; the odds ratio shows a larger numerical contrast than the relative risk (a pattern expected when the outcome is relatively common), and all measures point to a consistent direction and statistically significant magnitude of disparity in sentencing outcomes.

\textit{What ethical considerations are necessary when interpreting this result?}\\
Observed disparities in sentencing outcomes should be interpreted as evidence of differences in how sentences are imposed rather than as evidence of inherent differences in criminal propensity among ethnic groups. Explanations for disparities should emphasize systemic factors, such as prosecutorial discretion, implicit bias, differential charging practices, and structural inequalities, rather than characteristics of the groups themselves. Care must be taken to avoid attributing causation to race without further case-level analysis and to ensure that interpretations focus on sentencing disparities alone.

\textbf{Analytical Constraints and Considerations}\\
\textit{What is the input dataset unable to tell us and why?}
The input dataset does not by itself explain case-level drivers of sentencing decisions such as offense severity, plea bargaining details, prosecution charging rationales, judge-level discretion, or validated measures of prior criminal history; absence of these covariates limits ability to isolate whether observed disparities are attributable to race or to other case-specific factors. The data are anonymized and may lack certain variables excluded under disclosure rules, which prevents follow-up verification of individual case contexts and constrains the ability to control for potentially confounding factors. Statistical association observed in this aggregated data cannot establish causal attribution and may be influenced by measurement error, missing information, or systemic processes earlier in the criminal legal pipeline.

\textbf{Key Terms and Methodology}\\
The odds ratio is a ratio comparing the odds of an event occurring in one group to the odds in another group; it is commonly used to quantify relative differences in odds between two groups. Relative risk is the ratio of the probability of an event in one group to the probability in another group and conveys how much more or less likely the event is in one group compared with the other. The chi-square test is a statistical test that assesses whether two categorical variables are associated by comparing observed cell counts to expected counts under independence. A confidence interval is a range of values, derived from the data, that is believed to contain the true population parameter with a specified level of confidence. A p-value is the probability of observing data as extreme as those observed, or more so, if the null hypothesis of no association were true. The balance ratio is a descriptive measure indicating the relative sample sizes or proportions between comparison groups and is used to assess whether groups are similar in size for comparative analysis.

Read the full report here: \url{https://github.com/redoio/ai_evals/tree/main/report_001}

\end{quote}

\end{document}